\documentclass[conference, a4paper]{IEEEtran}
\IEEEoverridecommandlockouts
\usepackage{cite}
\usepackage{amsmath,amssymb,amsfonts, bm}
\usepackage[linesnumbered,vlined,ruled]{algorithm2e}
\usepackage{graphicx}
\usepackage{textcomp}
\usepackage{xcolor}
\usepackage{array, makecell}
\usepackage{multicol,multirow,booktabs,mathabx}
\def\BibTeX{{\rm B\kern-.05em{\sc i\kern-.025em b}\kern-.08em
    T\kern-.1667em\lower.7ex\hbox{E}\kern-.125emX}}

\newcommand{\funcfont}[1]{{\fontfamily{qcr}\selectfont\small $\thinspace$#1$\thinspace$}}
\newcommand{\nonl}{\renewcommand{\nl}{\let\nl\oldnl}}% Remove line number for one line
\let\oldnl\nl% Store \nl in \oldnl
\usepackage{setspace}

\usepackage[a4paper, left=1.6cm, right=1.6cm, top=1.9cm, bottom=4.5cm]{geometry}

\begin{document}

\title{3D Ground Truth Reconstruction from Multi-Camera Annotations Using UKF
}

\author{\IEEEauthorblockN{Linh Van Ma\IEEEauthorrefmark{1}, Unse Fatima\IEEEauthorrefmark{1}, Tepy Sokun Chriv\IEEEauthorrefmark{1}, Haroon Imran\IEEEauthorrefmark{1}, Moongu Jeon\IEEEauthorrefmark{1}}
\IEEEauthorblockA{\IEEEauthorrefmark{1}\textit{Department of Electrical Engineering and Computer Science, GIST, Korea}\\
\IEEEauthorrefmark{1}\{linh.mavan, unse.fatima, chrivsokuntepy, harooneecs, mgjeon\}@gm.gist.ac.kr}}

% \author{\IEEEauthorblockN{Linh Van Ma, Muhammad Ishfaq Hussain, JongHyun Park, Moongu Jeon}
% \IEEEauthorblockA{\textit{School of Electrical Engineering and Computer Science}, \\
% \textit{Gwangju Institute of Science and Technology, Republic of Korea}\\
% \{linh.mavan, ishfaqhussain, citizen135, nmgjeon\}@gist.ac.kr}
% }

\maketitle

\begin{abstract}
Accurate 3D ground truth estimation is critical for applications such as autonomous navigation, surveillance, and robotics. This paper introduces a novel method that uses an Unscented Kalman Filter (UKF) to fuse 2D bounding box or pose keypoint ground truth annotations from multiple calibrated cameras into accurate 3D ground truth. By leveraging human-annotated ground-truth 2D, our proposed method, a multi-camera single-object tracking algorithm, transforms 2D image coordinates into robust 3D world coordinates through homography-based projection and UKF-based fusion. Our proposed algorithm processes multi-view data to estimate object positions and shapes while effectively handling challenges such as occlusion. We evaluate our method on the CMC, Wildtrack, and Panoptic datasets, demonstrating high accuracy in 3D localization compared to the available 3D ground truth. Unlike existing approaches that provide only ground-plane information, our method also outputs the full 3D shape of each object. Additionally, the algorithm offers a scalable and fully automatic solution for multi-camera systems using only 2D image annotations.
\end{abstract}


\section{Introduction}

Three-dimensional (3D) multi-object tracking (MOT) has the potential to transform mobility and safety, such as autonomous driving, traffic control, or manufacturing. By accurately determining the position, orientation, and size of dynamic objects, such as pedestrians, cyclists, and vehicles, over time, 3D MOT enables safe and reliable navigation in complex environments.

To ensure precise annotations for 3D multi-object tracking, the authors \cite{chavdarova2018wildtrack,hou2020multiview} developed a time-efficient method using accurate camera calibration to position a 3D cylinder on a 12$\times$36m ground plane, discretized into a 480×1440 grid with 2.5cm resolution. The cylinder’s 2D projections form rectangles across seven camera views, aligned with annotated objects via a single adjustment, reducing the need for multiple bounding box annotations.  However, these methods only offer ground plane location, and consider an object as a point. Suppose two people share the same ground plane position but differ in height e.g., one standing on a balcony and the other on the first floor at the same 3D horizontal coordinates. In that case, they cannot be distinguished based solely on their ground plane position.
% Their Python-based web annotation tool, built with Django, displays synchronized images from all cameras, allowing users to place bounding boxes with a single click at an object’s base and adjust positions using keyboard controls. Real-time updates, a zooming feature, and an intuitive GUI enhance annotation accuracy and efficiency, with annotators improving precision over time, minimizing adjustments.

In the Panoptic Studio system \cite{Joo_2017_TPAMI}, 3D joint locations are estimated using multiview triangulation, which reconstructs objects' 3D positions and shapes by combining 2D keypoint detections from multiple calibrated camera views. Each camera captures 2D keypoints using a pose detector, and those with high confidence are selected for triangulation. By leveraging known camera parameters, the system finds the 3D point in space that best aligns with these 2D observations across views, minimizing reprojection error to ensure geometric consistency. The large number of synchronized cameras enables robust performance, allowing accurate, markerless 3D pose reconstruction even under occlusions and in complex human social interactions. Furthermore, their method is free from error accumulation, without suffering from motion jitter, and enables capture of long term group interactions (e.g., more than 10 minutes). Although this method can construct 3D pose ground truth but it only operates with points and cannot annotate with inputs that are bounding boxes.

\begin{figure}[ht]
\begin{centering}
\includegraphics[width=0.5\textwidth]{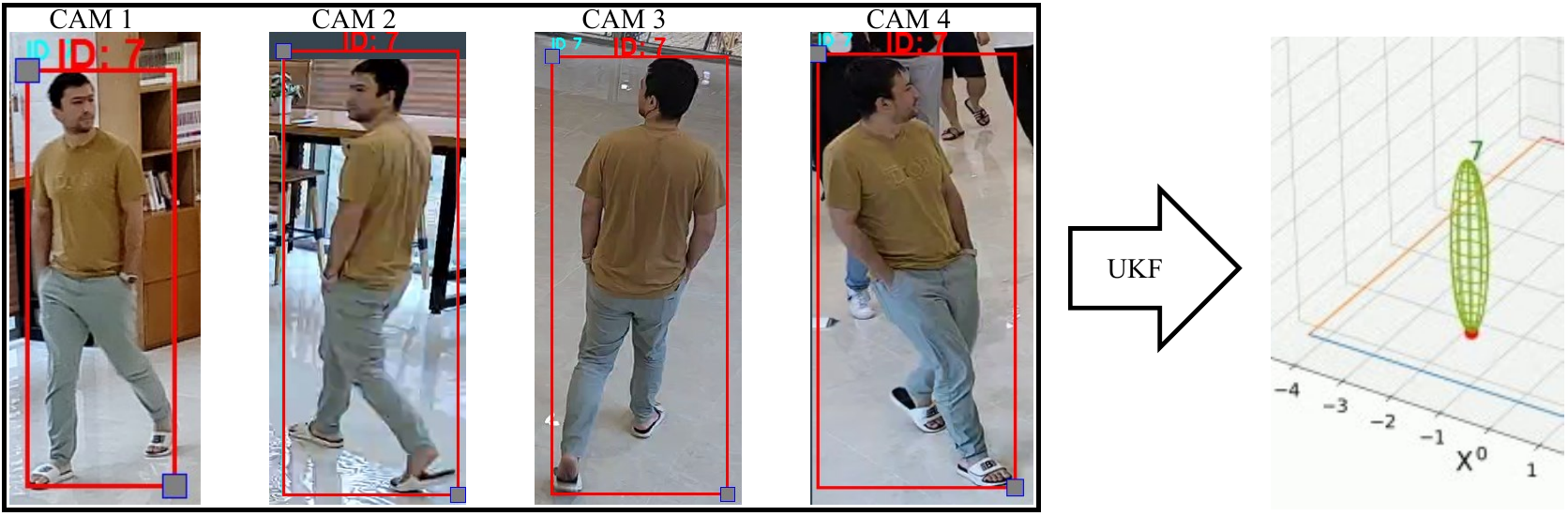}
\par\end{centering}
\caption{An illustration for the 3D annotation of our method. Our output is a 3D ellipsoid representing the height, width, and length of an object. In contrast, a big red point on the ground plane is the output of previous methods such as Wildtrack \cite{chavdarova2018wildtrack} and MultiviewX \cite{hou2020multiview}. \label{fig:fusecams}}
\end{figure}

In this research, we propose a 3D ground truth annotation method that leverages human-annotated 2D bounding boxes or pose keypoints from multiple cameras. For each object, we independently estimate 3D information using a single-object tracking algorithm. Our multi-camera single-object tracking algorithm transforms 2D image coordinates into robust 3D world coordinates using homography-based projection and unscented Kalman filter (UKF)-based fusion. The algorithm processes multi-view data to estimate object positions and shapes, addressing challenges like occlusions and varying camera perspectives. Fig. \ref{fig:fusecams} illustrates the difference between our annotation method and existing methods. Unlike prior methods such as Wildtrack, which represent each object as a single point on the ground plane (e.g., a red dot), our method outputs a full 3D ellipsoid capturing both position and shape. Evaluated on the Wildtrack and Panoptic datasets, our approach achieves high accuracy in 3D localization, even with sparse detections.

The paper is structured as follows. Section \ref{sec:backgrounds} reviews related work on 3D multi-camera multi-object tracking and multi-pose estimation, including ground truth annotation. Section \ref{sec:ann_algs} details the motion and measurement models for single-object tracking, together with the Unscented Kalman Filter (UKF), which are used to generate 3D shape and position annotations from 2D bounding boxes, and 3D pose annotations from 2D image-based pose ground truth. Section \ref{sec:experiment} provides experiments to validate the effectiveness of our proposed algorithms. Finally, Section \ref{sec:conclusion} concludes the paper.

\section{Related Works}\label{sec:backgrounds}

\noindent\textbf{3D Multi-Camera Multi-Object Tracking and Annotation}.\\
Recent approaches leverage deep learning models for 3D detection and tracking, such as deep-learning-based detectors and trackers \cite{chavdarova2017deep, baque2017deep}, spatial-temporal multi-cut formulations \cite{nguyen2022lmgp}, and ground plane information for early-fusion \cite{teepe2024earlybird}. However, these methods often focus only on ground plane locations, neglecting 3D shapes and poses. Some approaches, like \cite{gao2023delving}, produce 3D trajectories but rely on computationally intensive 3D detections, requiring retraining when camera geometry changes, limiting real-time applicability. Alternatively, filters like MS-GLMB \cite{ong2020bayesian, linh2024inffus, vo2019multi} using Labeled Random Finiset \cite{van2024visual}  to integrate camera geometry for direct 3D tracking from 2D detections, but their high computational cost makes them unsuitable for real-time tracking with many objects.

The most widely used dataset with an overlapping camera set-up is the PETS 2009 S2.L1 \cite{ferryman2009pets2009} sequence. In part due to the presence of a slope in the scene, the provided calibration poses large homography mapping deterioration and inconsistencies when projecting 3D points across the views. The three sequences shot at the EPFL campus \cite{fleuret2007multicamera}: Laboratory, Terrace, and Passageway, as well as SALSA \cite{alameda2015salsa} and Campus \cite{xu2016multi} are overlapping multi-camera datasets as well. The EPFL-RLC \cite{chavdarova2017deep} dataset demonstrates improved joint-calibration accuracy and synchronization compared to PETS. However, rather than providing a complete ground truth, this dataset represents a balanced collection of positive and negative multi-view annotations and is used for classifying a position as occupied by a pedestrian or not. Similar to WT \cite{chavdarova2018wildtrack} and MultiviewX \cite{hou2020multiview}, it uses triangulation to reconstruct 3D positions as single points on the ground plane, without providing object height, width, or length.

Similar to previous approaches, our annotation relies on the requirement of overlapping camera views. When an object is seen by only one camera, its 3D position remains highly uncertain along the projection ray. However, when multiple cameras observe the object, their complementary views reduce this uncertainty, resulting in a more accurate 3D state estimate shown in Fig. 1 \cite{van20243d}. Unlike existing methods, our approach outputs not only the 3D location but also the object's shape, specifically its width, length, and height.

\noindent\textbf{3D Multi-Object Pose Estimation and Annotation}.\\
Early 3D pose estimation modeled poses as graphical structures with joints as nodes and their relationships as edges, using maximum a posteriori (MAP) methods with 2D detections and physical constraints \cite{belagiannis20143d}. Modern methods often follow a two-step process: 3D human localization followed by detailed pose estimation. For example, VoxelPose \cite{tu2020voxelpose} uses two networks for localization and pose estimation, robust to varying camera views. TesseTrack \cite{reddy2021tessetrack} provides an end-to-end framework for 3D detection and temporal pose estimation. Graph neural networks \cite{wu2021graph} improves pose estimation using relational and projective attention techniques. SelfPose3D \cite{srivastav2024selfpose3d} generates 3D joints from 2D augmentations without 3D training data. Geometry-based, learning-free methods, such as plane sweep techniques \cite{lin2021multi} and cross-view correspondence \cite{liao2024multiple}, handle 3D reconstruction and generalize to new camera setups. %MVPose \cite{dong2019fast} uses 2D detection and Re-ID features for pose association, while \cite{chen2020multi} addresses occlusion issues by matching human feet across frames for MAP-based joint estimation.
In the Panoptic Studio system \cite{Joo_2017_TPAMI}, 3D joint locations are estimated using multiview triangulation, which combines 2D keypoint detections from multiple calibrated camera views to reconstruct their 3D positions. In contrast, we employ an Unscented Kalman Filter (UKF) to fuse all 3D pose keypoints. The UKF not only integrates multiview observations but also smooths keypoint trajectories without relying on additional smoothing techniques, effectively reducing motion jitter.

\section{Unscented Kalman Filter for Single Object Tracking from Multi-camera}\label{sec:ann_algs}
The single state attribute $x=\{\xi, \dot{\xi}, \sigma, p, \dot{p}\}$ of an object consists of 3D position $\xi$, 3D velocity $\dot{\xi}$, shape parameter $\sigma$, and 3D keypoints $p$ with its 3D velocity $\dot{p}$.
The shape parameter $\sigma$ is a triplet of (logarithms of) the half-lengths of the principal axes of the ellipsoid containing the object, and follows a random-walk model. The kinematics ($\xi$, $\dot{\xi}$, $p$, $\dot{p}$) follows a nearly constant velocity model, where $p$ is a list of 3D keypoints in 3D Cartesian coordinates. The number of keypoints varies depending on the chosen keypoint model, such as the COCO model with 17 keypoints \cite{belagiannis20143d} or the Panoptic model with 15 keypoints \cite{Joo_2017_TPAMI}. Here, we assume $\xi$, $\sigma$, and $p$ are independent, without accounting for the object’s body-structure constraints. Hence, the 3D keypoints are reconstructed independently from the 3D object representation derived from image bounding boxes, although their initialization is partially influenced by the bounding boxes. Below, we focus solely on the bounding box fusion process for constructing a 3D ellipsoid; the reconstruction of 3D keypoints follows the same procedure.

% \begin{table}[!ht]
% \centering
% % \global\long\def\arraystretch{1.3}%
%  \caption{List of symbols.\label{tbl:notationlist}}
%  %
%  \footnotesize
% \begin{tabular}{|c|>{\arraybackslash}p{6.5cm}|}
% \hline 
% Notation & Description\tabularnewline
% \hline 
% $b^{(k,c)}_t$ & 2D bounding box for the $k$-th object in the $c$-th camera view at time $t$\tabularnewline
% $v$ & Number of cameras\tabularnewline
% $\mu_t^{(k)}$ & Mean state vector of a target $t$ at time step $k$\tabularnewline
% $\mu_t^{(k,c)}$ & Updated mean state vector of a target $t$ at time step $k$ and camera $c$\tabularnewline
% $P_t^{(k)}$ & Covariance of a target $t$ at time step $k$\tabularnewline
% $P_t^{(k,c)}$ & Updated covariance of a target $t$ at time step $k$ and camera $c$\tabularnewline
% \hline 
% \end{tabular}
% \end{table}

% \subsection{Object Dynamic Model}\label{subsec:MOT-transition}

Our algorithm takes 2D bounding box ground truth annotations, denoted as \(\{b^{(k,c)}_t\}\), and estimates 3D positions and shapes of multiple objects over a sequence of time steps \(k = 1, 2, \dots, K\). The input annotation \(b^{(k,c)}_t\) represents the 2D bounding box for the \(t\)-th object in the \(c\)-th camera view at time step \(k\), where \(t = 1, 2, \dots, T\) indexes the objects, and \(c = 1, 2, \dots, v\) indexes the camera views. The output is a set \(\{\mathcal{T}\}\) containing the 3D position, shape, and keypoint estimates for all tracked objects across all time steps. %The algorithm leverages a combination of Kalman filtering for prediction and an Unscented Kalman Filter (UKF) for updating state estimates, ensuring robust tracking in a 3D environment.

\begin{figure*}[ht]
\begin{centering}
\includegraphics[width=0.8\textwidth]{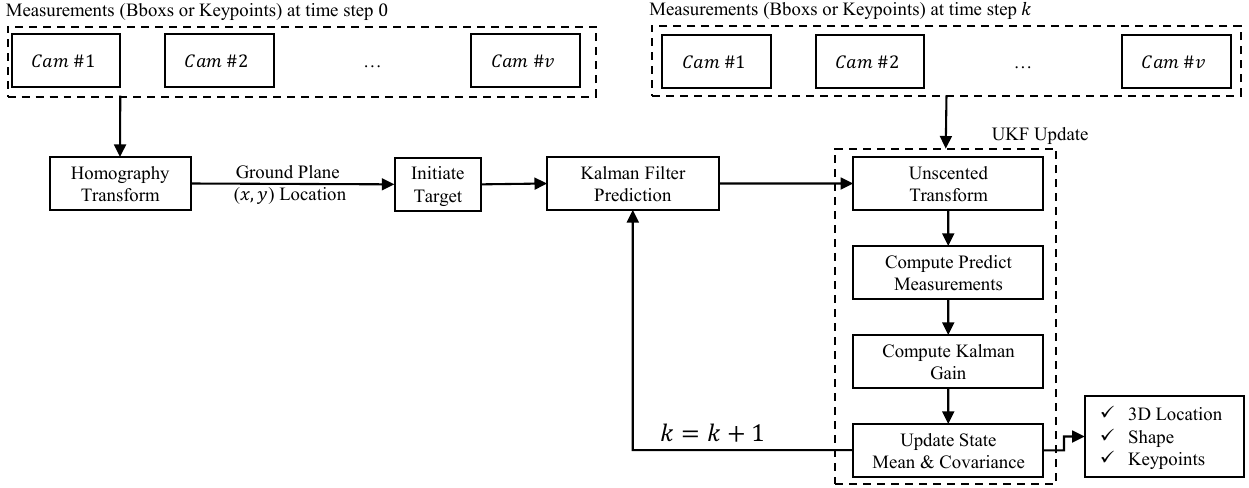}
\par\end{centering}
\caption{Single object tracking using an Unscented Kalman Filter (UKF) to estimate 3D location, shape, and keypoints from measurements, including bounding boxes and keypoints, obtained from multiple camera images. We assume a target born at time step $0$, it survives and evaluates from time step $k$ to $k+1$. \label{fig:singleukf}}
\end{figure*}

 An overview of our algorithm is presented in Fig. \ref{fig:singleukf} and Alg. \ref{alg:singleukf}. Each step is described in detail in the following subsections. For each object \(t\), the algorithm begins with track initialization, followed by prediction and state update using measurements (bounding boxes and keypoints), and concludes with state extraction to estimate the object’s geometric representation.

\begin{algorithm}%%%%%%% Overall Algo %%%%%%%%
\SetInd{0.4em}{0.5em}
\setstretch{0.9}
\fontsize{10pt}{10pt}
\DontPrintSemicolon
\nonl \textbf{Input}: \nonl 2D bounding box ground truth annotations $\left\{b^{(k,c)}_t\right\}$. \;
\nonl \textbf{Output}: \{$\mathcal{T}$\} 3D Positions {\&} Shapes.
\rule[0.5pt]{0.925\columnwidth}{0.1pt} \;
\For{$t=1:T$}{
\funcfont{/*Track Initialization*/} \;
$\left(\mu_t^{(0)}, \mathbf{P}_{t}^{(0)}\right)$$\leftarrow$ \funcfont{InitTarget$\left(b^{(0,c)}_t\right)$} // \cite[Alg. 1]{linh2024inffus}\;

\For{$k=1:K$}{
    \funcfont{/*Track Prediction*/}\;
    $\left(\mu_t^{(k)}, \mathbf{P}_t^{(k)}\right)$$\leftarrow$\funcfont{KalmanPrediction}$\left(\mu_t^{(k)}, \mathbf{P}_t^{(k)}\right)$ \;
    
    \funcfont{/*Track Update*/} \;
    $j = b_t^{(k)}$ //\funcfont{BBox index of object $t^{th}$}\;
    \For{$c=1:v$}{
        \If{$j^{(c)}>0$}{
           $\left(\mu_t^{(k,c)},\mathbf{P}_{t}^{(k,c)}\right)\leftarrow $\funcfont{UKFUpdate}$\left(b_t^{(k,c)},\mu_t^{(k)}, \mathbf{P}_{t}^{(k)}\right)$ \cite[Alg. A1]{linh2025tmm} or \cite[Alg. 7]{ma2025visual}\;
           $\mu_t^{(k)}=\mu_t^{(k,c)}; \mathbf{P}_{t}^{(k)}=\mathbf{P}_{t}^{(k,c)}$\;
        }
    }
    $\mathcal{A} \leftarrow$ \funcfont{Extract3DEstimates}$\left(\mu_t^{(k)}\right)$\;
    $\mathcal{T}^{(t)} \leftarrow$ \funcfont{Append}($\mathcal{T}^{(t)}, \mathcal{A}$) 
}
$\mathcal{T} \leftarrow$ \funcfont{Append}($\mathcal{T}, \mathcal{T}^{(t)}$)
}
\caption{Algorithm iteration at time $t$. \label{alg:singleukf}}
\end{algorithm}

% \begin{algorithm}
% \SetInd{0.4em}{0.5em}
% \setstretch{0.9}
% \fontsize{10pt}{10pt}
% \DontPrintSemicolon
% \nonl \textbf{Input}: Mean $\mu_t^{(k)}$ \& Covariance $\mathbf{P}_{t}^{(k)}$, measurement model $h(\cdot)$, measurement noise $R$, detected bounding boxes $b_t^{(k,c)}$. \;
% \nonl \textbf{Output}: mean $\mu_t^{(k,c)}$, covariance $\mathbf{P}_{t}^{(k,c)}$ at camera $c$.
% \rule[0.5pt]{0.925\columnwidth}{0.1pt} \;
% \funcfont{// (1) Generate $s$ Sigma Points \& Weights} \;
% $\left(\left\{\mathcal{X}^{i}\right\}_{i=1}^s, \left\{w^i\right\}_{i=1}^s\right) \leftarrow$ \funcfont{UT} $\left(\mu_t^{(k)}, \mathbf{P}_{t}^{(k)}\right)$ \;
% \funcfont{// (2) Measurements Prediction} \;
% \For{$i = 1:s$}{
%     $\mathcal{Y} ^i \leftarrow h\left(\mathcal{X}^i\right)$
% }
% $\hat{x} =\sum_{i=1}^s w^i \mathcal{X}^i$\;

% \funcfont{// (3) Compute $\mu_t^{(k)}, \mathbf{P}_{t}^{(k)}$ with measurements} \;
% $\hat{y} \leftarrow \sum_{i=1}^s w^i \mathcal{Y} ^i$ \;
% $S  \leftarrow R  + \sum_{i=1}^s w^i \left(\mathcal{Y} ^i - \hat{y}\right)\left(\mathcal{Y} ^i - \hat{y}\right)^T$ \;
% $\Psi  \leftarrow \sum_{i=1}^s w^i \left(\mathcal{X} ^i - \hat{x}\right)\left(\mathcal{Y} ^i - \hat{y}\right)^T$ \;
% $\mu_t^{(k,c)} \leftarrow \mu_t^{(k)} + \Psi  S ^{-1} (b_t^{(k,c)}  - \hat{y})$ \;
% $\mathbf{P}_{t}^{(k,c)} \leftarrow \mathbf{P}_{t}^{(k)} - \Psi  S ^{-1} \Psi ^T$ \;

% \caption{Unscented Kalman Filter Update}
% \label{alg:ukf}
% \end{algorithm}

\subsection{Track Initialization}
We initialize a new object (referred to as a target) in a 3D multi-camera single object tracking system by estimating its initial 3D position based on 2D bounding box ground truth annotations from multiple cameras. The process begins by iterating over the number of cameras, $v$. For each camera \( c \), we extracts the ground truth bounding boxes for the specified target. For valid ground truth bounding boxes, we compute the 2D feet location as the midpoint of the bounding box’s bottom edge. This 2D point is transformed into a 3D ground plane position using a homogeneous transformation with the inverse of the camera parameters. The 3D ground plane positions from all sensors are then averaged along the horizontal dimensions to compute the mean of the object’s ground plane position. A state vector is initialized with its position (using the average location ) and velocity components. To initialize multiple objects simultaneously, a mean shift algorithm, described in \cite[Alg. 1]{linh2024inffus}, is employed to ensure that each object’s track is established from a reliable starting point based on 2D image observations. However, in this work, we only need to initialize a single target at a time.

To initialize the 3D keypoints of a person, we begin with a standardized canonical human pose, represented by a fixed set of keypoint coordinates in a normalized space. This standard pose is then aligned with the estimated object center from the state vector by translating it to match the mean position of the target. To ensure the keypoints fit within the spatial extent of the estimated ellipsoid representing the target's body, we scale the keypoints in the vertical (y) and height (z) dimensions according to the ellipsoid’s corresponding axes. Additionally, each keypoint is augmented with velocity components, replicated from the target's overall velocity, to form a full kinematic state vector for filtering. %This initialization provides a consistent and geometry-aware basis for incorporating pose information into subsequent tracking and filtering steps.

\subsection{Track Prediction and Update}
The algorithm then iterates over \(K\) tracking steps for each object. In the track prediction, we employ a Kalman filter. This step advances the state estimate \(\mu_t^{(k-1)}\) and covariance \(\mathbf{P}_t^{(k-1)}\) from the previous step to the current time \(k\) using a state transition model. The state transition model assumes a linear motion model for object $t$ at time step $k$, defined as:
\begin{equation}
\mu_t^{(k)} = \mathbf{F} \mu_{t}^{(k-1)} + \mathbf{w},
\end{equation}
where \(\mathbf{F}\) is the state transition matrix (modeling constant velocity motion), and \(\mathbf{w}\) is mean Gaussian process noise with covariance \(\mathbf{Q}\). For a 3D position and velocity state vector \(\mu = [\xi, \dot{\xi}, \sigma]^T\), the transition matrix \(\mathbf{F}\) is:
\begin{equation}
\mathbf{F}=\left[\begin{array}{cc}
I_{3}(\Delta t) & 0_{6\times3}\\
0_{3\times6} & I_{3}
\end{array}\right],I_{3}(\Delta t)=I_{3}\otimes\left[\begin{array}{cc}
1 & \Delta t\\
0 & 1
\end{array}\right],
\end{equation}
where \(\Delta t\) is the sampling period. The covariance is updated as:
\begin{equation}
\mathbf{P}_t^{(k)} = \mathbf{F} \mathbf{P}_{t}^{(k-1)} \mathbf{F}^T + \mathbf{Q}.
\end{equation}
This prediction accounts for the expected motion of the object while incorporating uncertainty due to process noise.

\begin{figure}[ht]
\begin{centering}
\includegraphics[width=0.5\textwidth]{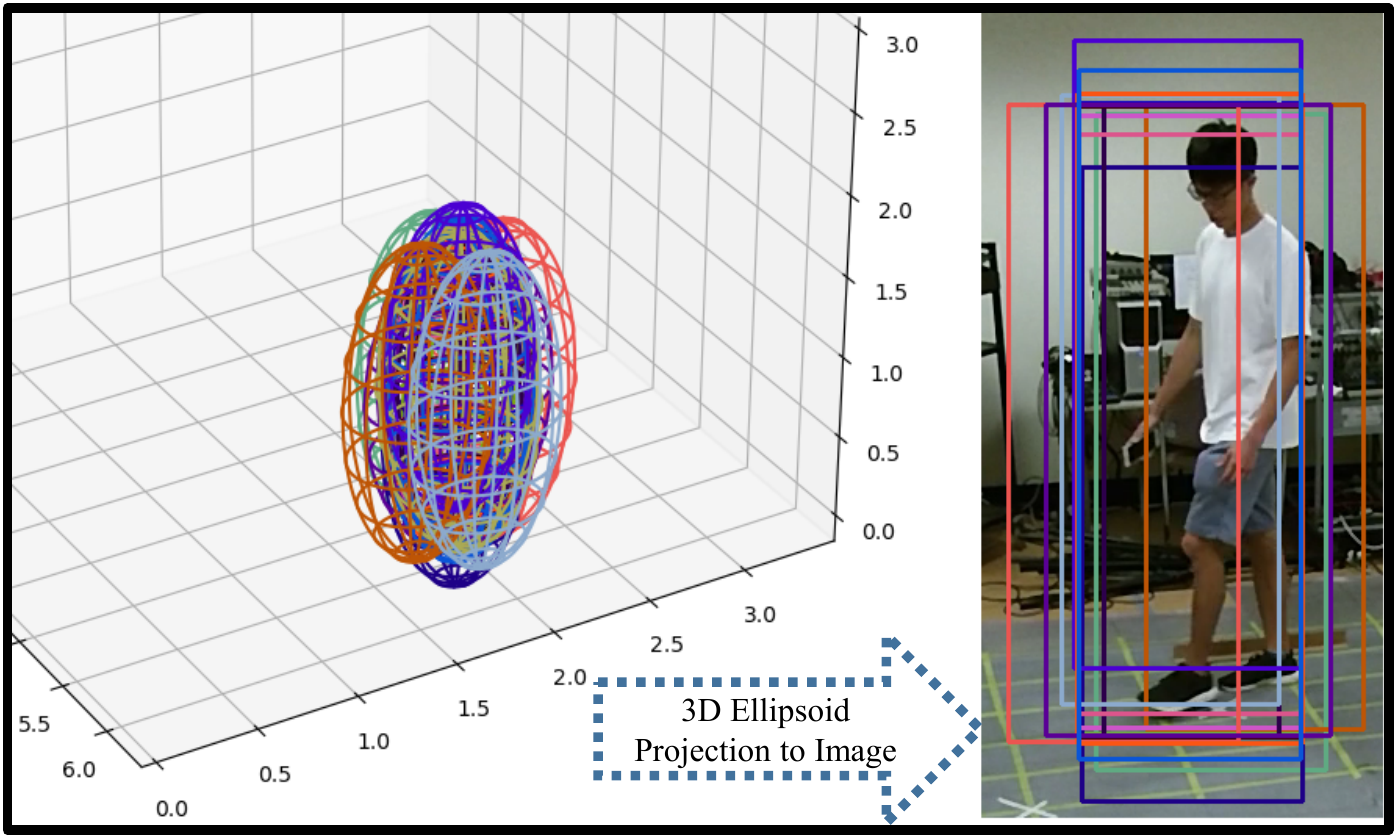}
\par\end{centering}
\caption{We represent an object as an ellipsoid and employ the unscented transform to generate $(2d) + 1$ ellipsoids (\cite{van2004sigma,julier2004unscented}), capturing the object's uncertainty in 3D space, where $d$ denotes the dimension of the object state. Each ellipsoid is then projected onto a 2D image as a bounding box using a camera matrix. \label{fig:ukfprojected2bbox}}
\end{figure}

In the track update phase, the algorithm processes observations from each camera view \(c\). For each object \(t\) at \(k\) time step, the bounding box index \(j = b_t^{(k)}\) identifies the corresponding 2D bounding box of the object $t$. If a valid bounding box exists in view \(c\), (\(j^{(c)} > 0\) indicates that the object is observed in the \(c\)-th camera view, while \(j^{(c)} = 0\) means that the object is not observed in this view), we update its state by using the UKF update, detailed in \cite{julier2004unscented} or \cite[Alg. A1 \& Alg. A4]{linh2025tmm}. We adopt the UKF because it accommodates nonlinear projections between 3D space and 2D camera planes. Specifically, the measurement model relates the state \(\mu_t^{(k,c)}\) to the 2D bounding box \(b_t^{(k,c)}\) via a nonlinear function:
\begin{equation}
z_t^{(k,c)} = h(\mu_t^{(k,c)}) + \mathbf{v},
\end{equation}
where \(z_t^{(k,c)}\) is the observed 2D bounding box, \(h(\cdot)\) is the measurement function, and \(\mathbf{v}\) is zero-mean Gaussian measurement noise with covariance \(\mathbf{R}\). Fig. \ref{fig:ukfprojected2bbox} illustrates how the Unscented Transform projects a 3D ellipsoid onto the image plane for measurement prediction, which constitutes part of the UKF update step for estimating the mean $\mu_t^{(k)}$ and covariance $\mathbf{P}_t^{(k)}$. Specifically, the function \(h(\cdot)\) generates a set of sigma points around the predicted state \(\mu_t^{(k,c)}\), which are then projected onto the camera plane to obtain predicted bounding boxes. These predicted measurements are used to computed the Kalman gain. The implementation of \(h(\cdot)\) can be realized through camera projection using quadric projection \cite[pp. 201]{hartley2003multiple} or approximated using the approach in \cite[Alg. A6]{linh2025tmm}. The state and covariance are then updated using the Kalman gain, aligning the estimates with the observed 2D bounding box.

After updating the state for all views, we extract 3D estimates \(\mathcal{A}\) from the updated state \(\mu_t^{(k)}\). These estimates include the object’s 3D position, shape parameters, and pose keypoints. The estimates \(\mathcal{A}\) are appended to the track \(\mathcal{T}^{(t)}\) for the current time step. At the end of each time step, the track \(\mathcal{T}^{(t)}\) is appended to the global set \(\mathcal{T}\), which accumulates the 3D tracking results across all time steps.

\section{Experiments}\label{sec:experiment}

\noindent \textbf{Dataset and Evaluation Metrics}: In this section, we evaluate our algorithm's performance on multi-camera multi-object tracking datasets, including Curtin Multi-Camera (CMC) \cite{ong2020bayesian}, WILDTRACK (WT) \cite{chavdarova2018wildtrack}, and MultiviewX \cite{hou2020multiview}, as well as 3D pose estimation datasets, CMU Panoptic (CMU) \cite{joo2015panoptic}. % including Campus, Shelf \cite{belagiannis20143d}, and 

\begin{table*}[h!]
\centering{}
\global\long\def\arraystretch{1.2}%
 \caption{3D ground truth annotation performance of our algorithm on WT and MultiviewX datasets for the entire sequence of 400 frames. CMC dataset contains five sequences, the evaluation is performed for the entrire sequences.\label{tbl:3dposshape}}
 \scriptsize
\begin{tabular}{|c|cccccc|}
\hline 
Sequence & FP$\downarrow$ & FN$\downarrow$ & IDs$\downarrow$ & MOTA$\uparrow$ & IDF1$\uparrow$ & OSPA\!$^{\text{(2)}}$\!$\downarrow$ \tabularnewline
\hline 
WT \cite{chavdarova2018wildtrack} & 61 & 61 & 3 & 98.7 & 99.3 & 0.1\tabularnewline
\hline 
MultiviewX \cite{hou2020multiview} &  269 & 269 & 2 & 96.5 & 98.3 & 0.17\tabularnewline
\hline 
CMC1 \cite{ong2020bayesian} & 0 & 0 & 0 & 99.5 & 99.8 & 0.01 \tabularnewline
\hline 
CMC2 \cite{ong2020bayesian} & 2 & 2 & 0 & 99.9 & 99.4 & 0.05 \tabularnewline
\hline 
CMC3 \cite{ong2020bayesian} & 1 & 3 & 0 & 99.7 & 98.9 & 0.04 \tabularnewline
\hline 
CMC4 \cite{ong2020bayesian} & 0 & 0 & 0 & 99 & 99 & 0.02 \tabularnewline
\hline 
CMC5 \cite{ong2020bayesian} & 7 & 0 & 0 & 99.8 & 99.8 & 0.0 \tabularnewline
\hline 
\end{tabular}
\end{table*}

\begin{table*}
    \centering
    \caption{\small{Average Precision (AP) and Metrics for 3D Pose Estimation at Various Distances (AP@25 to AP@150), Recall at 500mm, Mean Per Joint Position Error (MPJPE) at 500mm}} % evaluation code using Faster-VoxelPose/lib/dataset/panoptic.py
    \label{tab:panoptic}
     \scriptsize
        \begin{tabular}{l|cccccc}
            \toprule
             PanOptic Seqs                                                             & AP$_{25}$     & AP$_{50}$     & AP$_{100}$    & AP$_{150}$    & Recall$_{@500}$ & MPJPE[mm]     \\
            \midrule
            160906\_pizza1 &  97.7 & 99.7 & 99.8 & 99.8 & 99.8 & 8.2 \\
            \cmidrule{1-7}
            160906\_ian5 &  77.6 & 98.6 & 99.7 & 99.7 & 99.7 & 14.7 \\
            \midrule
            160906\_band4 &  99.0 & 99.0 & 99.2 & 99.1 & 99.2 & 6.1 \\
            % tt_id_start = np.amin(gt_bboxes[gt_bboxes[:, 2]== tt_id, 0]) + 25
            \cmidrule{1-7}
            160422\_haggling1 &  88.2 & 97.9 & 98.9 & 99.1 & 99.2 & 9.1 \\
            % tt_id_start = np.amin(gt_bboxes[gt_bboxes[:, 2]== tt_id, 0]) + 15
            \bottomrule
        \end{tabular}
    % }
\end{table*}

For WILDTRACK and MultiviewX, we generate 3D ground truth from bounding box annotations. For CMU Panoptic, we back-project 3D ground truth keypoints to the image plane using the camera calibration matrix and reconstruct 3D keypoints, evaluating with MPJPE for qualitative comparison. Tracking performance is evaluated using CLEAR MOT metrics \cite{bernardin2008evaluating} (MOTA, MT, ML, FP, FN, IDS), IDF1 \cite{ristani2016performance}, HOTA \cite{luiten2021hota}, and OSPA$^{\texttt{(2)}}$ \cite{beard2020asolution,rezatofighi2020trustworthy}, which measures tracking error (lower is better). %For 3D keypoint estimation, we assess performance on the CMU dataset. We use the PCP3D metric \cite{dong2019fast} to measure the percentage of correctly estimated body parts and an extended Average Precision ($AP_K$) \cite{pishchulin2016deepcut}, where a pose is correct if its Mean Per Joint Position Error (MPJPE) \cite{tu2020voxelpose} is below a threshold ($K$ mm).

\noindent \textbf{Parameter Settings}: We set the covariance noise and measurement noise to be ignored since we using the ground truth. The evaluation threshold is 1 meter for the Euclidean distance. Similarly, the cut-off distance for the OSPA$^{\texttt{(2)}}$ metric is set to 1 meter for Euclidean base-distance. In our implementation, the process noise covariance $\mathbf{Q}$, mean Gaussian noise process $\mathbf{w}$, and measurement noise covariance $\mathbf{R}$ are configured to approach zero, reflecting the assumption that input measurements/detections represent ground truth.
%For Campus and Shelf, we construct 3D keypoints directly from image-based ground truth keypoints and use the percentage of correctly estimated parts for qualitative assessment.

\noindent \textbf{Evaluation Results}: On the CMC dataset, we achieve nearly perfect performance. It suggests that CMC and our method has a similar structure on 3D ground truth data annotation. On the WT dataset, Tab. \ref{tbl:3dposshape}, our method achieved a MOTA of 98.7 and an IDF1 score of 99.3, with only 61 false positives, 61 false negatives, and 3 identity switches, indicating precise object localization and strong identity consistency. Similarly, on the MultiviewX dataset, the method maintained high accuracy with a MOTA of 96.5 and an IDF1 of 98.3, while keeping the number of identity switches as low as 2. The OSPA$^{\texttt{(2)}}$ distances, 0.10 for WT and 0.17 for MultiviewX, further confirm the spatial closeness between our predicted 3D locations and the ground truth. Additionally, we evaluated our algorithm on multiple Panoptic Studio sequences to assess 3D pose estimation performance shown in Tab. \ref{tab:panoptic}. Across four sequences \textit{\{160906\_pizza1, 160906\_ian5, 160906\_band4, 160422\_haggling1\}}, the results show consistently high average precision, with AP${25}$–AP${150}$ values exceeding 97\% in most cases, and recall at 500mm reaching up to 99.8\%. The MPJPE remains low, ranging from 6.1 mm to 14.7 mm, indicating precise joint localization. These results collectively demonstrate that our method offers high-fidelity 3D annotation and pose estimation that closely align with the ground truth across diverse multi-camera datasets. 

From our perspective, the slight discrepancies in performance falling just short of 100\% can be attributed to the initialization mechanism not being perfectly aligned during the 3D pose estimation process. This limitation presents an opportunity for improvement and will be explored in future research. For the WT and MultiviewX datasets, our method achieves near-perfect performance. The minor differences observed are primarily due to the inherent variation between our algorithm’s object center estimation approach and the ground truth center annotations provided in these datasets. It is also important to note that the ground truth data in these benchmarks may not fully represent the actual physical ground truth, as absolute accuracy would require costly manual measurement. In this context, we conclude that our annotations are comparable to the provided ground truth and closely approximate the real-world ground truth.

\begin{figure}[ht]
\begin{centering}
\includegraphics[width=0.5\textwidth]{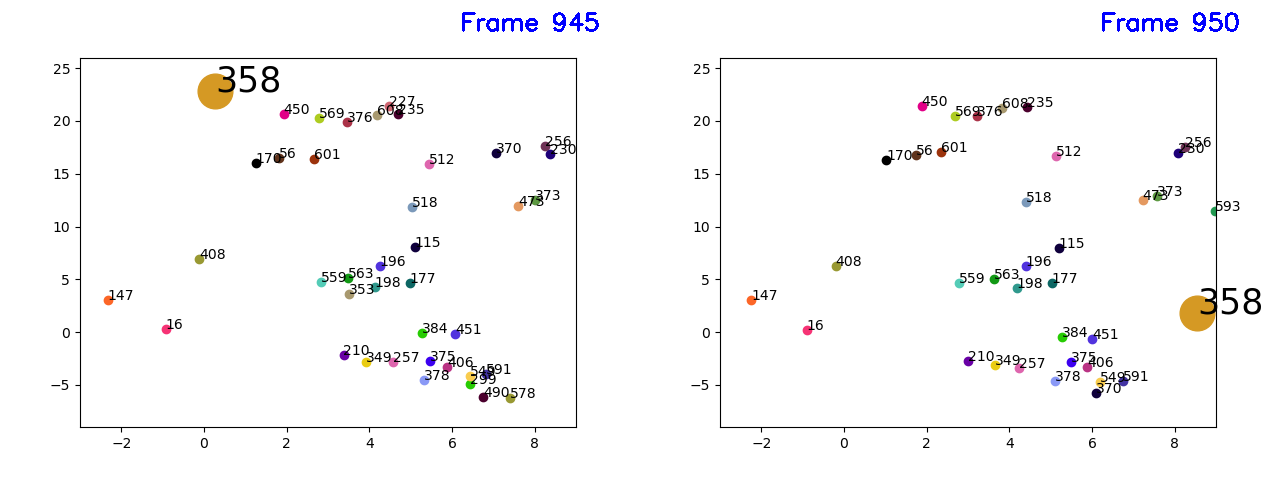}
\par\end{centering}
\caption{Illustration of the annotation error for object ID 358 in the Wildtrack dataset. The object disappears at frame 945 and reappears at frame 950 in a distant and inconsistent location, resulting in an implausible jump in Wildtrack dataset \cite{chavdarova2018wildtrack}. \label{fig:wt_wrong_id}}
\end{figure}

Furthermore, during qualitatively checking performance of our 3D annotation in the Wildtrack dataset, we observe that the ground truth annotation for object ID 358 is incorrect. The object exits the scene at frame 945 and reappears at frame 950 in an implausible location. The two positions are far apart and not aligned in direction, making it appear as if the object suddenly jumps backward. This discrepancy is illustrated in Fig. \ref{fig:wt_wrong_id}.

\section{Conclusion}\label{sec:conclusion}
This paper introduces a novel 3D ground truth annotation method that leverages 2D ground truth bounding boxes or pose keypoints from multiple camera views to generate accurate 3D object representations. Unlike prior approaches that reduce objects to points on the ground plane, our method estimates full 3D ellipsoids, capturing position, dimensions, and shape by fusing multi-view observations using a homography-based projection and an unscented Kalman filter. Experimental results on the CMC, Wildtrack, MultiviewX, and Panoptic datasets demonstrate that our method achieves high accuracy, closely approximating the ground truth provided by these datasets. By providing richer 3D annotations without relying on dense keypoint detection or extensive manual labeling, our approach broadens the applicability of multi-view tracking systems and enables more realistic modeling of real-world scenarios for autonomous systems and vision research. Nonetheless, our method heavily relies on overlapping fields of view, precise calibration, and accurate ground-truth annotations in all views, which opens room for future research.

\section*{Acknowledgements}
This work was supported by the National Research Foundation of Korea(NRF) grant funded by the Korea government(MSIT) (No. 2023R1A2C2006264).

\bibliographystyle{IEEEtran}
\bibliography{reflib}

\end{document}